\documentclass[conference,compsoc]{IEEEtran} 
\usepackage{graphicx}
\graphicspath{ {my-images/} }

\usepackage{hyperref}       
\usepackage{url}            
\usepackage{booktabs}       
\usepackage{amsfonts}       
\usepackage{nicefrac}       
\usepackage{microtype}      
\usepackage{xcolor}         
\usepackage{multirow}
\usepackage{multicol}

\usepackage{subfigure}

\usepackage{stfloats}
\usepackage{lipsum}

\title{Deepfake detection in videos with multiple faces using geometric-fakeness features}

\ifCLASSOPTIONcompsoc
\usepackage[nocompress]{cite}
\else
\usepackage{cite}
\fi

\usepackage{amsmath}
\usepackage{algorithmic}
\usepackage{array}
\usepackage{url}
\usepackage{comment}

\begin{document}

\author{Kirill Vyshegorodtsev, Dmitry Kudiyarov, Alexander Balashov, Alexander Kuzmin \\
Cybersecurity Laboratory, Sberbank of Russia \\
\texttt{ \shortstack{vke88@ya.com, dmitry.kudiyarov@gmail.com,\\ balashov.alvi@gmail.com, alex.kuzminx@gmail.com}}
}

	\maketitle

	\begin{abstract}
		Due to the development of facial manipulation techniques in recent years deepfake detection in video stream became an important problem for face biometrics, brand monitoring or online video conferencing solutions. In case of a biometric authentication, if you replace a real datastream with a deepfake, you can bypass a liveness detection system. Using a deepfake in a video conference, you can penetrate into a private meeting. Deepfakes of victims or public figures can also be used by fraudsters for blackmailing, extorsion and financial fraud. Therefore, the task of detecting deepfakes is relevant to ensuring privacy and security. In existing approaches to a deepfake detection their performance deteriorates when multiple faces are present in a video simultaneously or when there are other objects erroneously classified as faces. In our research we propose to use geometric-fakeness features (GFF) that characterize a dynamic degree of a face presence in a video and its per-frame deepfake scores. To analyze temporal inconsistencies in GFFs between the frames we train a complex deep learning model that outputs a final deepfake prediction. We employ our approach to analyze videos with multiple faces that are simultaneously present in a video. Such videos often occur in practice e.g., in an online video conference. In this case, real faces appearing in a frame together with a deepfake face will significantly affect a deepfake detection and our approach allows to counter this problem. Through extensive experiments we demonstrate that our approach outperforms current state-of-the-art methods on popular benchmark datasets such as FaceForensics++, DFDC, Celeb-DF and WildDeepFake. The proposed approach remains accurate when trained to detect multiple different deepfake generation techniques.		
	\end{abstract}
	
	\section{Introduction}
	Just a few years ago deepfakes were used as an enter-tainment content without any claims to realism. However, the technology has developed further and quickly became a useful tool for realistic video manipulations and was even weaponized for performing cyberattacks and committing fraud. Freely available deepfake generation tools [4], [11], [31], [43] have been developed and the resulting videos can deceive many internet users. Those tools have already allowed fraudsters in some countries to commit successful fraud, bypassing a facial biometric authentication and using deepfakes of known public figures in fraudulent schemes. In the fall of 2021 several Eastern European countries were targeted by a fraud campaign with a deepfake video of a famous banker offering to invest in his new financial project and promising co-financing from his bank [10], [42]. Especially dangerous are fraudsters from Russia. In March 2021 several European politicians fell victim to deepfake video calls impersonating Russian political activists [38], [54]. Other public figures, celebrities, or top managers of large companies around the world may soon also become unwilling actors in a deepfake fraud. However, the threats of deepfakes are not limited to reputational losses for companies or individuals. Facial biometric authentication systems are also threatened by deepfakes. Such systems are widely used in the financial sector all around the world allowing a convenient method of authentication for many clients [23], [27]. Fraudsters can intercept a video stream from the client’s device camera, replacing a real video with a deepfake and successfully authenticating in an app or a service. Similar approach has been used for deepfake video calls on Zoom [3], [4]. Even liveness detection algorithms used in many facial biometric systems are not able to protect against deepfakes. The method proposed in this article offers a single solution able to detect multiple different deepfake generation techniques with a state-of-the-art accuracy. From an ethical standpoint we pursue a goal of protecting the media space and people on the Internet from a new form of a digital identity theft that might incur reputational, financial, and other losses.

	\section{Related Work}
	A continuous development of image generation and manip¬ulation methods allowed creating very realistic fake images and videos. And their efficient and accurate detection becomes nowadays more and more relevant. Let us review previous research done in this area. Early works on deepfake detection methods investigated artifacts in face images [1], [6], [24], [46]. They used deep neural network models such as ResNet [22], SE-ResNet [25], EfficientNet [52] etc. The output of those models was fed into a fully connected network that performed an image classification. Other works investigate different statistical properties of colors in an image, a spectral decomposition and color spaces [15], [41], [44], [65]. In more recent works models base their predictions on position of different artefacts in an image [2], [5], [26], [48], [51], [58].   
	
	Next generation of works focus on the analysis of sequences of frames rather than single images, considering a temporal dimension. Some works focus on certain areas of a face in an image to detect deepfakes. For example, Li et al base their method on the detection of eye blinking that is not well presented in synthesized fake videos. Whereas another work uses a spatiotemporal network pretrained on a lip-reading task to detect unnatural lip movements [21]. Chen et al. propose to use Multi-scale Patch Similarity Module (MPSM), which measures a similarity between features of local regions and forms a robust and generalized similarity pattern [8]. Further works study different inconsistencies to locate traces of forgery in deepFake videos, e.g., S-MIL [35] and STIL [17]. However, ignoring some parts of a face may lead to a lower deepfake detection accuracy. Mittal et al [40] use Siamese network to analyze similarities between audio and visual modalities of the same video, extracting and comparing affective cues corresponding to perceived emotions in both modalities. Whereas Sabir et al. [47] exploit temporal information in a video by using a recurrent convolutional model. The researchers propose to use a system made up of a convolutional, a recurrent and a feed forward neural network to combine information from multiple frames. Good results were also demonstrated by a combination of convolutional networks and a LSTM network [28], [29], [33]. Tariq, S., Lee, S. and Woo, S have proposed a similar architecture CLRNet - Convolutional LSTM-based Residual Network [53].  
	
	As we can see in several works training a single model to detect multiple deepfake generation techniques leads to a significant decrease in accuracy [1], [30], [46]. So, development of a generalizable deepfake detection method becomes a relevant task. Du et al. [14] propose to detect deepfakes using a locality-aware autoencoder. Li and Lyu [36] propose an approach based on detecting face warping artifacts caused by a difference in a resolution between a deepfake region and an original image. An approach that employs expanding the training sample and generating additional images is proposed in the works by Xueyu Wang et al. [57], Liang Chen et al. [9], Kaede Shiohara et al. [50] at the CVPR conference in 2022. A novel idea that yielded a significant increase in accuracy is proposed by Li et al in their work Face X-ray [34]. They analyze whether an input image can be decomposed into a blending of two images from different sources and detect a blending boundary between those two images. Their approach is shown to be effective when applied to deepfakes generated by unseen face manipulation techniques. The work of Khan and Dai [30] employs a video transformer for a deepfake detection. They use an incremental learning on different datasets to train a generalizable model. In addition to F2F, FS, NT and DF [46] they use DeepFake Detection (DFD) [16] and Deepfake Detection Challenge (DFDC) [12] datasets for their training. The use of transformers has also been researched in a work by Xiaoyi Dong [13].  
	
	In recent works by Zhihao Gu, Yang Chen, Taiping Yao et al. presented at AAAI-2022 and IJCAI-2022 [18], [19] the authors proposed a novel sampling unit named a snippet which contains several successive video frames for learning local temporal inconsistencies that can serve as a strong indicator of deepfakes. Whereas previously a deepfake detection was performed on sparsely sampled video frames and therefore did not take into consideration such local information. Moreover, they have designed an Intra-Snippet Inconsistency Module (Intra-SIM) and an Inter-Snipper Interaction Module (Inter-SIM), where Intra-SIM is meant to mine short-term motions within each snippet and Inter-SIM is devised to promote a cross-snippet information interaction to form global representations. Such modules work in an alternate manner and can be plugged into existing 2D CNNs. Their approach proved to be successful and outperformed existing state of the art competitors on popular datasets, such as FaceForensics++, DFD, Celeb-DF, DFDC, WildDeepFake. New improvement methods have been developed by J. Liang's team. The original method of improvement efficiency using masking is given in the research [60]. A simple yet effective strategy named Thumbnail Layout (TALL) and its improvement TALL++ is given in the researchs [61], [62]. The algorithm transforms a video clip into a pre-defined layout to realize the preservation of spatial and temporal dependencies.    
	  
	\textbf{Ethics}. Eliminating bias and discrimination by demographic, ethnic and other attributes is an important part of our research. Therefore, we have used datasets with representatives from many different groups:  
	
	- In terms of geography: Africa, Asia, North America, South America, Australia, Europe, Middle East, South Asia and other areas.  
	
	- In terms of ethnicity: Afro-Americans, Asian, Indian, Arabs, African, Native American, European, Latinos, Multiracial and others.  
	
	- In terms of other attributes: different age, gender, height, weight, hair texture, skin color and other attributes are also represented in the datasets.  
	
	This helps us eliminate or at least minimize any discrimination or bias issues.

	\section{Method}
	\subsection{Motivation}
	Deepfake generation is a hot topic of research nowadays, so new, more advanced, and realistic deepfake generation algorithms are being released quite often. This poses a challenge for a deepfake detection since the models trained to detect a particular deepfake method do not perform well on previously unseen algorithms. If a separate model is trained for each new deepfake generation algorithm, such an ensemble would quickly become computationally inefficient due to a large and ever-growing number of separate models required. If a single model is trained to detect different deepfake generation methods, the accuracy of their detection drops significantly due to a variability in artifacts introduced by different deepfake generation methods. A quick and accurate inference response is especially relevant to a task of biometric authentication and identification since modern biometric systems must handle many requests simultaneously. A potential attacker may employ a camera virtualization and a live deepfake generation software in a single pipeline to pass active liveness checks (e.g., blinking or head movement) with a deepfake. Such instruments are freely available online as an opensource software and do not require extensive knowledge to be used [55]. And even if biometric software implements protection against using virtual cameras, such a mechanism is easy to bypass. Therefore, the main motivation of our research is to propose a single model able to detect multiple deepfake generation methods without significantly sacrificing its accuracy. Additionally, such a model must possess a good generalizing ability to successfully detect previously unseen deepfake generation methods.

	\begin{figure}[!htbp]
		\centering
		\includegraphics[width=.85\linewidth]{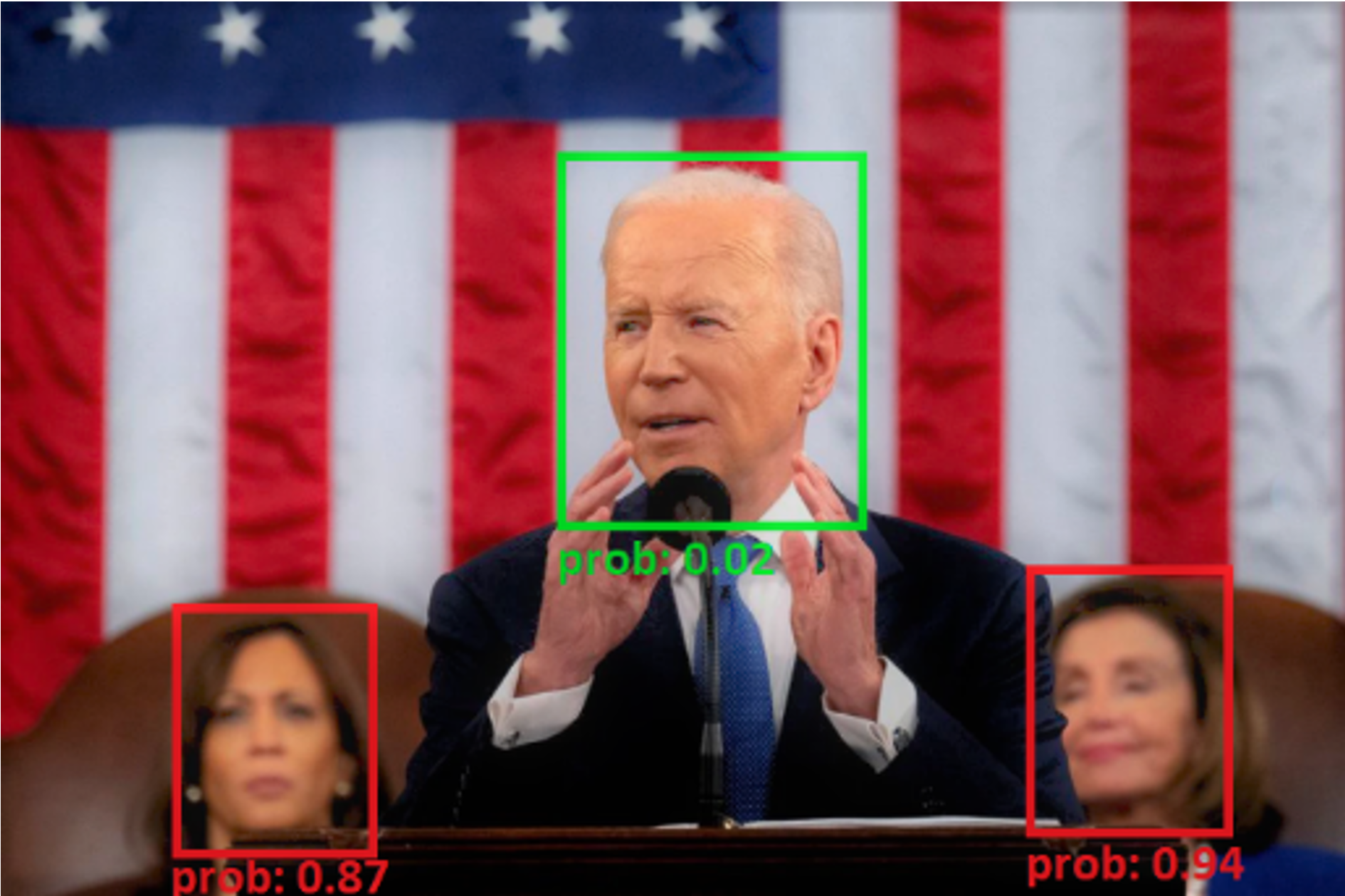}
		\caption{Example of false positives of the model on background faces [45].}
		\label{fig:mesh1}
	\end{figure}

	Another challenging problem is a simultaneous presence of multiple faces in a video. Portraits and other background faces often cause a model to erroneously classify a valid video as a deepfake thus lowering the detection accuracy (Fig.~\ref{fig:mesh1}).

\begin{figure}[!htbp]
		\centering
		\begin{minipage}[h]{0.99\linewidth}
			\center
			\subfigure[]{\includegraphics[width=.85\linewidth]{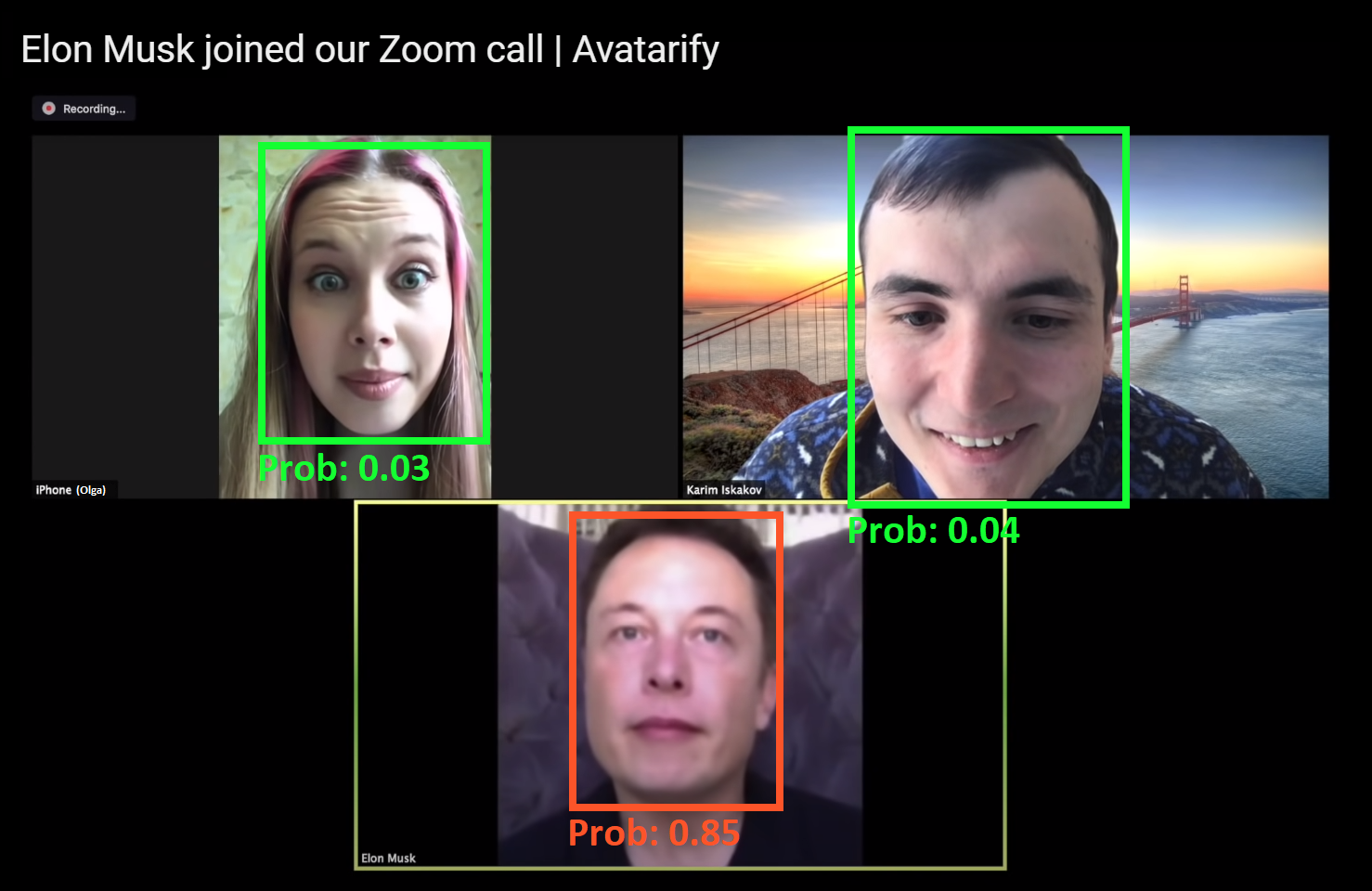} \label{fig:Fig_2_zoom} }
			
		\end{minipage}
		\vfill
		\begin{minipage}[h]{0.99\linewidth}
			\center
			\subfigure[]{\includegraphics[width=.85\linewidth]{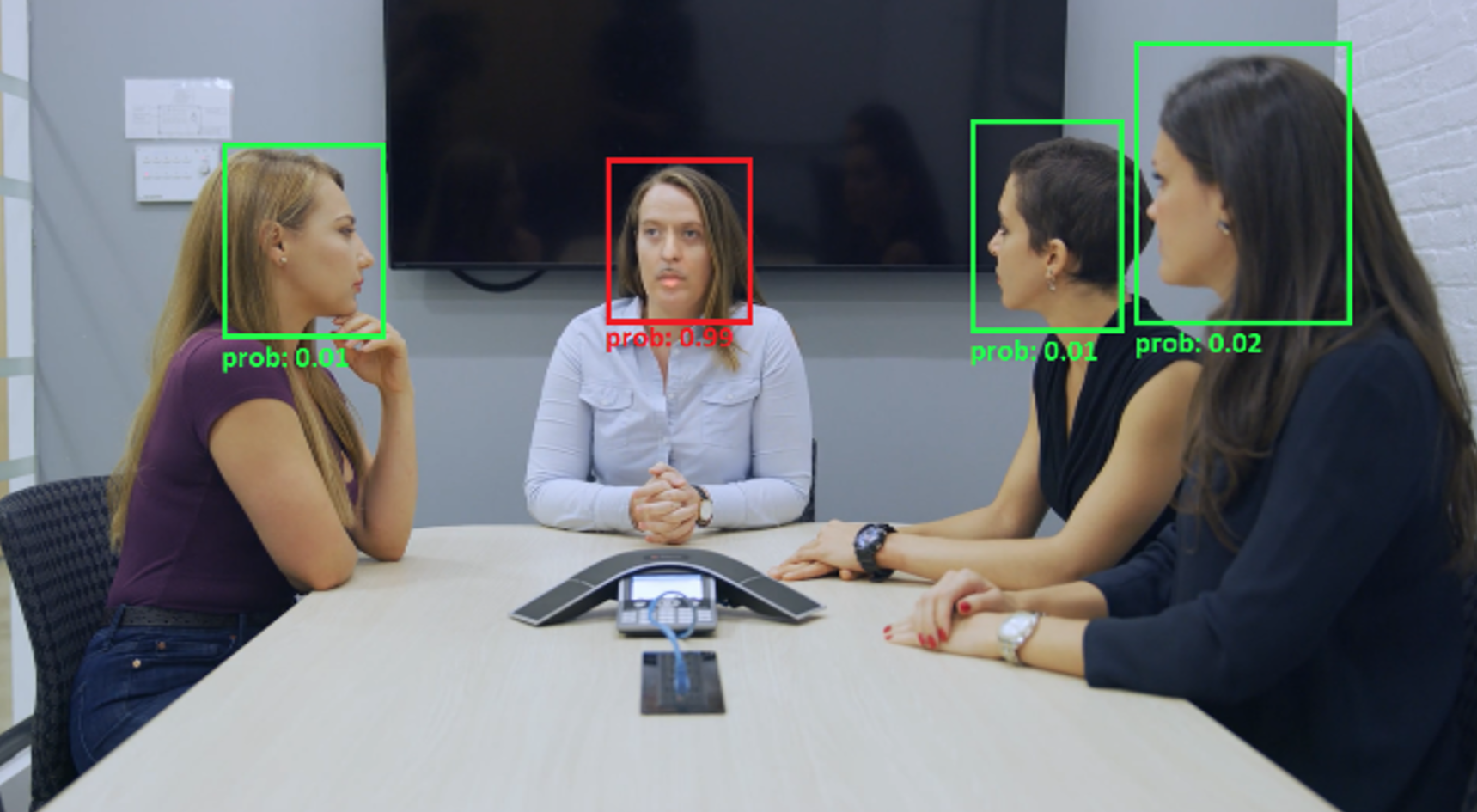} \label{fig:Fig_2}}
		\end{minipage}
		\caption{Examples of the presence of real faces and deepfakes. Averaging the scores for all the faces in the video computed a low score. \subref{fig:Fig_2_zoom} Deepfake in meeting in Zoom [3]; \subref{fig:Fig_2} Meeting at the conference (video from DFD dataset [16])}
		\label{fig:mesh2}
	\end{figure}

	At the same time, one cannot simply dismiss faces smaller than a certain threshold value or all non-primary faces in the same video, since a secondary face can still significant enough to be targeted by a deepfake attack (Fig.~\ref{fig:mesh2}). By placing several real faces in a video together with a deepfake a potential attacker can significantly affect the final prediction for a video passing a deepfake video for a real one. This problem is crucial for deepfake detection in brand monitoring and online video conferencing solutions where the presence of multiple faces in a video is very common. This problem is especially relevant to ensuring security and privacy of online video conferences and meetings. An attacker can create a deepfake of a top manager or a government official and use it in a private video call or a conference. Such attacks are quite easy to perform using the same camera virtualization and live deepfake generation software as we have mentioned before. A proof-of-concept of an attack has already been demonstrated with Zoom video calls (Fig.~\ref{fig:Fig_2_zoom}). Other real faces in the same video stream with a deepfake (e.g., attacker's accomplices) will seriously hinder deepfake detection in this case. Figure 2b shows an example of a proposed attack (a deepfake is clearly visible for demonstration purposes).	
	Therefore, in our research we would like to address those problems and propose a suitable and computationally efficient solution.

	\subsection{Developing geometric-fakeness features of faces for deepfake detection}
	Our model consists of several different blocks of neural networks that we train together. The first block is based on the EfficientNet-B4 architecture [52] (denoted as DNNBlock in Figs.~\ref{fig:mesh5}-~\ref{fig:mesh6}). It computes the characteristics of a fake -fakeness features. We hypothesized that erroneous per frame predictions are random in their nature whereas reliable scores have stable distribution patterns. A score of an individual frame has a relation to the scores of previous frames. To model this relation, our model includes blocks of convolutional neural network (denoted as CNNBlock in Figure~\ref{fig:mesh6}). While researching real and fake videos, we have discovered a connection between false positives and a location of a face in a frame. Indeed, secondary background faces that are unlikely to be faked occupy smaller area in a frame and appear on fewer frames than primary faces. We decided to formalize this information and use it in training our model further. To achieve this, we analyze a video and group together faces of the same person on different frames. Then we compute new features for each person in a video called geometric features, and use them to train our neural network. Our algorithm contains following steps.   
	
	\textbf{In Step 1} we send faces to the deep learning models (DNNBlock in Figs.~\ref{fig:mesh5}-~\ref{fig:mesh6}). Those models produce fakeness features - per frame output for each face in a video by each DNNBlock. 
	
	\textbf{In Step 2}, the faces of the same person in different frames are grouped together. To achieve this a pre-trained FaceNet model [49] is employed. Any other model that allows to calculate the distance between vectors corresponding to a measure of similarity between faces can be used instead. A vector is calculated for each face in each frame of a video. Then the frames of the video are processed sequentially to measure the distance between face vectors in adjacent frames. If the distance between vectors is below a preset threshold value, it is assumed that the faces belong to the same person and are therefore grouped together. For subsequent measurements of distance between vectors we take the weighted moving average (WMA) of vectors belonging to the same face in previous frames. The number of frames ($T$) used for calculating a WMA-vector of a face was set to 1/10 of a total number of frames in a video.
	\medskip 
	\[
	\overline{e}_{i,t+1} = \alpha\cdot e_{i,t+1} + \left(1-\alpha \right) \cdot \frac{ \sum_{f=1,T}^{} \left(1-\alpha \right)^{f} \cdot \overline{e}_{i,t+1-f}}{\sum_{f=1,T}^{} \left(1-\alpha \right)^{f}}
	\]
	\medskip 
	Where $\overline{e}_{i,t+1}$ is a WMA-vector of the i-th face on the $(t+1)$-th frame. This was performed to adapt the face vector to changes in a head position or lightning conditions. At the end of Step 2 the fakeness features from Step 1 for each face are aggregated into an ordered sequence represented as a vector - Fakeness features for separate faces (Figure~\ref{fig:mesh5}). This vector shows how fakeness features for each person in a video change in time.  
	
	\textbf{In Step 3}, geometric characteristics of each face in a video are produced and a relative area of a frame occupied by a face is calculated. This operation is performed for each face in each frame of a video. A total area occupied by all faces in the frame is also considered in the geometric characteristics to account for the relative importance of each face in a frame and to reduce an influence of a face that is less significant for a potential attacker and therefore is less likely to be faked. In a case where there are small faces of passers-by in the background and several larger faces in a foreground, e.g., an interview on a street, those smaller faces are probably random people and would not be targeted by an attacker. However, in a case where there are many smaller faces without bigger ones, e.g., a shooting in a hall of a theatre, those faces might be more significant and therefore more likely to be targeted by an attacker. Thus, multiplication by the total area of all faces is performed in order to distinguish between those two cases. As a result, geometric characteristics for each face are calculated using a formula below:
	\medskip 
	\[
	g_{i,t} = \frac{w_{i,t} \cdot h_{i,t} }{ w_{frame,t} \cdot h_{frame,t}} \sum_{j=1,F}^{} \frac{w_{j,t} \cdot h_{j,t} }{ w_{frame,t} \cdot h_{frame,t}}
	\]
	\medskip 
	Where $g_{i,t}$ – is a geometric characteristics of the $i$-th face ($h_{i,t}$ is height and $w_{i,t}$ is width of the face) on the $t$-th frame, where there are a total of $F_{t}$ faces. If a certain face is not present in a certain frame, its geometric characteristic is set to 0. At the end of step 3 for each person in a video we receive geometric features (Figure~\ref{fig:mesh5}) of his face in a frame. After that for each person, we combine this vector and the fakeness features vector from step 1 into a matrix.  
	
\begin{figure}[!htbp]
		\centering
		\includegraphics[width=.95\linewidth]{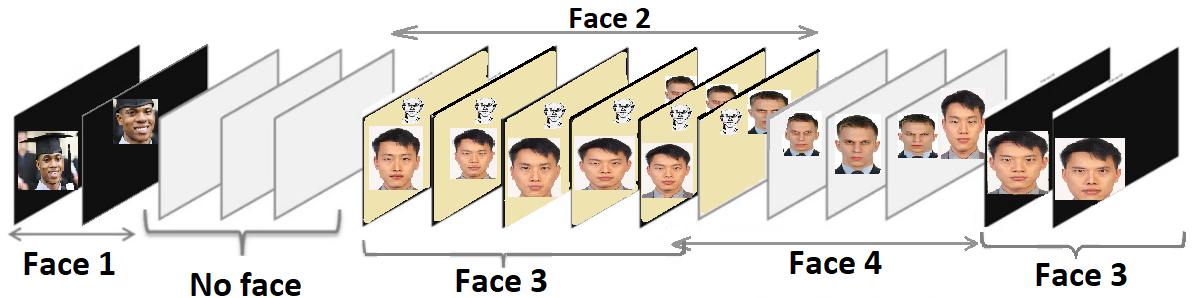}
		\caption{An example of the distribution of faces in a video.}
		\label{fig:mesh3}
	\end{figure}

	We called this matrix geometric-fakeness features (GFF) of faces. An example of creating such a GFF matrix for 16 frames of a video is given in Figure~\ref{fig:mesh3}.
	
\begin{figure}[!htbp] 
		\centering
		\includegraphics[width=.90\linewidth]{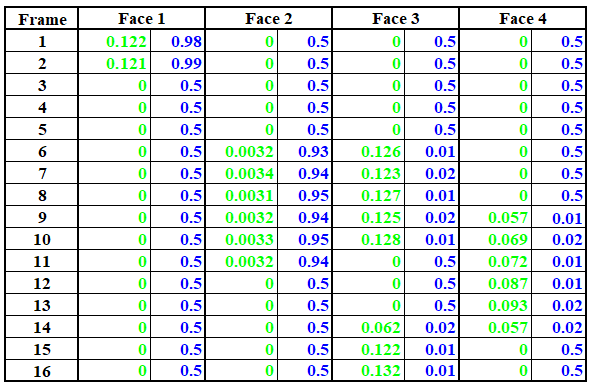}	
		\caption{Geometric-fakeness features for all faces in a video. For each face the first column (marked green) represents a sequence of per-frame geometric features (i.e. a relative area of a frame occupied by this face) and the second column (marked blue) represents a sequence of fakeness features (a detached prediction of the model that this face is a deepfake).}
		\label{fig:mesh4}
	\end{figure}

\begin{figure*}[!htbp]	
		\centering
		\includegraphics[width=.90\linewidth]{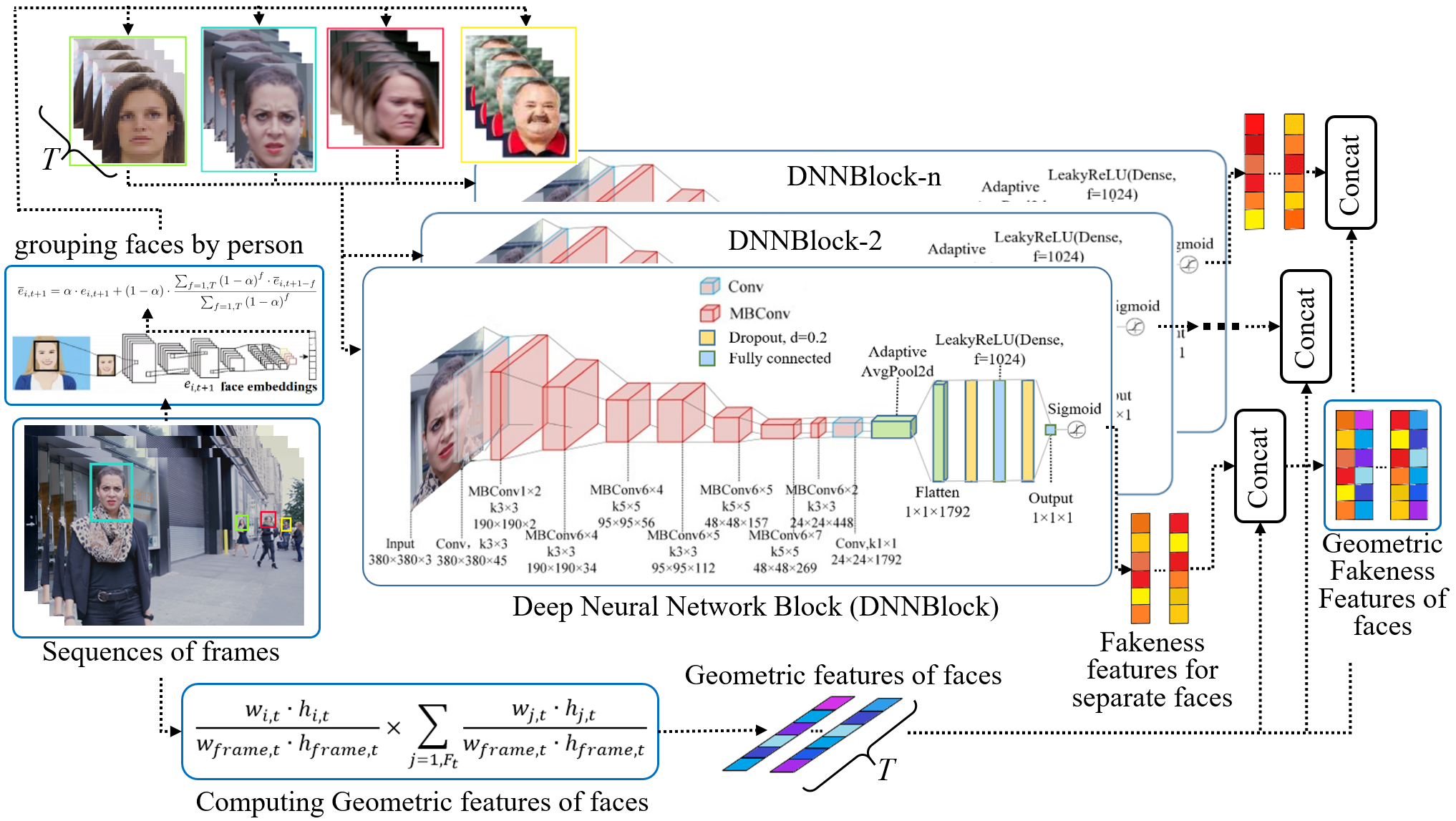}
		\caption{Architecture for computing geometric-fakeness features of faces in a video. Faces on a video are detected, embeddings of faces are computed to group the faces on different frames by person. To obtain fakeness features the faces are fed into DNNBlocks. Geometric features are calculated using the provided formula and are concatenated with fakeness features to form a GFF matrix.}
		\label{fig:mesh5}		
	\end{figure*}

\begin{figure}[!htbp]  
		\centering
		\includegraphics[width=8.0cm]{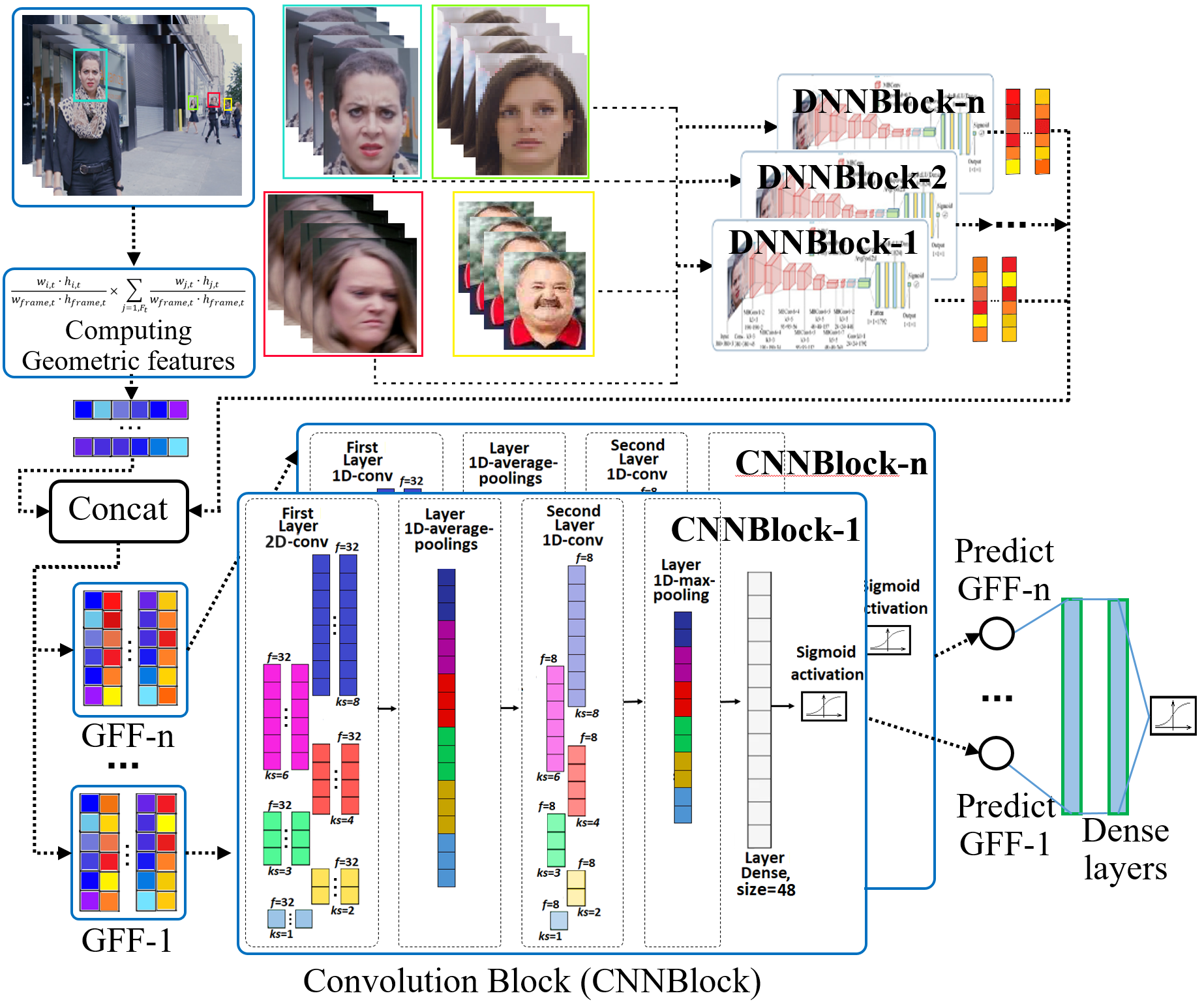}
		\caption{CNNBlock architecture for processing geometric-fakeness features. GFFs are processed by a CCNBlock with two convolution layers. A sequence of predictions outputted by the CNNBlock is formed into a vector that is processed by a fully connected neural network for a final prediction for a video.}
		\label{fig:mesh6}
	\end{figure}

	In this example faces 3 and 4 are real, face 1 is a random person passing by and face 2 is a portrait on a wall in the background. Computed geometric-fakeness features are given in Figure~\ref{fig:mesh4}. First column for each face (marked green) contains the area size of a face, second column (marked blue) – fakeness features for a given face changing in time. In this example, face 1 is incorrectly classified as a deepfake. This would have negatively impacted a final deepfake prediction for a video if existing analysis methods were used. However, in our approach, due to its limited temporal presence in a video we can deduce that it belongs to a passerby. Face 2 (the portrait) is also incorrectly classified as a deepfake. But using its geometric characteristic we can see that it is a background face that is unlikely to be faked.

	The complete process of creating GFF is shown in Figure~\ref{fig:mesh5}. Concatenated geometric-fakeness features of 5 faces are formed. Within the matrix, we sort the faces by their average fakeness value. If there are less than 5 faces in the video, GFF is filled with the default values. If there are more than 5 faces in the video, we form groups of 5 faces and feed them to the model. A maximum score among such groups used as a final prediction for the video.

	Next, the computed GFFs are passed into the CNNBlock of our model. The CNNBlock (see Figure~\ref{fig:mesh6}) is a two-layer CNN with 2D convolution in the first convolution layer and 1D convolution in the second convolution layer. A set of filters of different sizes is used in each layer. The best result was achieved by an architecture with 32 filters with kernel sizes 1, 2, 3, 4, 6, 8 and an average pooling layer. The second convolution layer used 8 filters with the same kernel sizes followed by a max-pooling layer. Output from max-pooling was fed into a dense layer of size 48, followed by a neuron with a sigmoid activation function. The CNNBlock forms prediction for each GFF. A sequence of outputs from the CNNBlock is formed into a vector.  
	
	Both DNNBlock and CNNBlock are trained together as a single model. During training images of faces are fed into the DNNBlock. The output of DNNBlock is concatenated into a tensor of fakeness features. Then the fakeness features are combined with computed geometric features into a new tensor and fed into the CNNBlock. The output of the CNNBlock is compared with the label for the video and is used it to obtain losses. This loss is propagated back through CNNBlock  and then DNNBlock. The final prediction for a video is computed using a fully connected neural network with 2 layers and a sigmoid output.

	\section{Experiments}
	\subsection{Experimental settings}
	\textbf{Datasets.} We evaluate our approach on five well-known and widely used benchmark datasets: FaceForensics++ [46], DFDC [12], Celeb-DF [37], DFD [16] and WildDeepFake [64].
	
	•	\textbf{FaceForensics++} is a forensics dataset consisting of 1000 original video sequences and 4000 videos (1000 videos for each method) that have been manipulated with four automated face manipulation methods: Deepfakes (DF), Face2Face (F2F), FaceSwap (FS) and NeuralTextures (1000 videos each). The videos are of different quality e.g., high quality (HQ) with almost no visual loss and low quality (LQ), which are visually blurry [46]. 
	
	•	\textbf{Deepfake detection challenge dataset (DFDC)} is a dataset consisting of 5000 videos featuring two facial modification algorithms. Diversity in several axes (gender, skin-tone, age, etc.) has been considered and actors recorded videos with arbitrary backgrounds thus bringing visual variability [12].
	
	•	\textbf{Celeb-DF} is a dataset of 590 real videos and 5639 high quality deepfake videos of celebrities. The real source videos are based on publicly available YouTube video clips of 59 celebrities of diverse genders, ages, and ethnic groups. The deepfake videos are generated using an improved deepfake synthesis method. As a result, the overall visual quality of the synthesized deepfake videos in Celeb-DF is greatly improved when compared to existing datasets, with significantly fewer notable visual artefacts [37].
	
	•	\textbf{The Deep Fakes Dataset (DFD)} is a dataset provided by Google \& Jigsaw and contains over 3000 manipulated videos from 28 actors in various scenes. The videos in the dataset are diverse samples in terms of actors gender, age, ethnic groups and in term of illumination, motion, pose, cosmetics, occlusion, content, and context [16].
	
	•	\textbf{WildDeepFake} is a dataset which consists of 7314 face sequences extracted from 707 deepfake videos collected completely from the internet. Compared to existing virtual deepfake datasets, WildDeepFake contains more diverse scenes, more persons in each scene and rich facial expressions. WildDeepFake is known for being a challenging dataset, where the detection performance can decrease drastically [67].
	
	We did not use OpenForensics [32] or similar image-based datasets, since they consist of separate non-related images whereas our approach is aimed at detecting deepfakes in a video (i.e. a sequence of related frames).. Therefore, OpenForensics cannot be used for our task. Our experiments were performed on the same datasets that are used by the research community for detecting deepfakes in videos (see Tables~\ref{table:table1}---~\ref{table:table4}).
	
	\textbf{Implementation}. We use a Multi-task CNN (MTCNN) for a face area detection [63] and a pre-trained EffIcientNet B4 to train a DNNBlock. The computed fakeness features for separate faces are concatenated with the geometric features of faces to form GFFs. They are fed into a CNNBlock. Predictions for each GFF are concatenated into a tensor. This tensor of GFF predictions is fed into a fully connected network. We used 5 DNNBlocks. The training was done using a binary cross-entropy (BCELoss) as a loss function and SGD with Nesterov as an optimizer (parameters: lr=0.001, momentum=0.9, batch size=12). A size of a cropped face is set to 380x380 pixels. We also used label smoothing equal to 0.001. The training was performed on a system with Intel Xeon E5-2680 CPU and 4 Nvidia Tesla T4 GPU. The model was trained for 100 epochs with 1000 samples each.

	\begin{table*}[!htbp]
		\caption{Comparison with state-of-the-art methods on FaceSwap, Deepfakes, Face2Face, NeuralTextures, DFD and DFDC dataset (results are presented in terms of F-measure).}
		\label{table:table1}
		\centering
		\begin{tabular}{@{}llllllll@{}} 
			\toprule
			\shortstack[l]{Method} & \shortstack[l]{FS} & \shortstack[l]{DF} & \shortstack[l]{F2F} & \shortstack[l]{NT} & \shortstack[l]{DFD} & \shortstack[l]{DFDC} & \shortstack[l]{Average\\(without DFDC)} \\
			\midrule
			\shortstack[l]Conv-LSTM & 0.5745 & 0.6312 & 0.5523 & 0.8012 & 0.5554 & - & 0.6229\\
			Xception & 0.7329 & 0.7857 & 0.7417 & 0.8245 & 0.5435 & - & 0.7257 \\
			XNet-ViT & 0.8640 & 0.8564 & 0.8371 & 0.5895 & 0.8624 & 0.8540 & 0.8019 \\
			MesoNet & 0.8017 & 0.8421 & 0.8252 & 0.8760 & 0.8627 & - & 0.8415\\
			CLRNet & 0.8720 & 0.8572 & 0.8920 & 0.8928 & 0.8652 & - & 0.8758\\
			\textbf{Ours (GFF)} & \textbf{0.9235} & \textbf{0.9288} & \textbf{0.9045} & \textbf{0.8945} & \textbf{0.9158} & \textbf{0.9103} & \textbf{0.9110}\\
			\bottomrule
		\end{tabular}
	\end{table*}
	
	\begin{table*}[!htbp]
		\caption{Comparison with the state-of-the-art deepfake detection methods on FF++ dataset (results are presented in terms of accuracy).}
		\label{table:table2}
		\centering
		\begin{tabular}{@{}lllllllll@{}}
			\toprule
			\addlinespace[4pt]
			\multirow{2}{50pt}[-1em]{Method} & \multicolumn{4}{c}{FaceForensics++ HQ} &  \multicolumn{4}{c}{FaceForensics++ LQ}\\
			\cmidrule(r){2-5} \cmidrule(){6-9} \\
			{} & DF & F2F & FS & NT & DF & F2F & FS & NT \\
			\addlinespace[4pt]
			\cmidrule(r){1-1} \cmidrule(r){2-5} \cmidrule(){6-9} \\
			
			Xception & 0.9893 & 0.9893 & 0.9964 & 0.9500 & 0.9678 & 0.9107 & 0.9464 & 0.8714 \\
			TEI & 0.9786 & 0.9714 & 0.9750 & 0.9429 & 0.9500 & 0.9107 & 0.9464 & 0.9036 \\
			DSANet & 0.9929 & 0.9929 & 0.9964 & 0.9571 & 0.9679 & 0.9321 & 0.9536 & 0.9178 \\
			V4D & 0.9964 & 0.9929 & 0.9964 & 0.9607 & 0.9786 & 0.9357 & 0.9536 & 0.9250 \\
			Co-motion-70 & 0.9910 & 0.9325 & 0.9830 & 0.9045 & - & - & - & -\\
			S-MIL & 0.9857 & 0.9929 & 0.9929 & 0.9571 & 0.9679 & 0.9143 & 0.9464 & 0.8857\\
			S-MIL-T & 0.9964 & 0.9964 & \textbf{1.0} & 0.9429 & 0.9714 & 0.9107 & 0.9607 & 0.8679\\
			STIL & 0.9964 & 0.9929 & \textbf{1.0} & 0.9536 & 0.9821 & 0.9214 & 0.9714 & 0.9178\\
			I-I-SIM & \textbf{1.0} & 0.9929 & \textbf{1.0} & 0.9643 & \textbf{0.9928} & 0.9571 & 0.9786 & \textbf{0.9428}\\
			\textbf{Ours (GFF)} & \textbf{1.0} & \textbf{0.9973} & \textbf{1.0} & \textbf{0.9854} & 0.9893 & \textbf{0.9762} & \textbf{0.9813} & 0.8935 \\
			\bottomrule
		\end{tabular}
	\end{table*}

	\textbf{Baselines}. We compare our method to other state-of-the-art methods. We mainly considered methods, that are aimed at Increasing a generalizability of a model to new and unseen image manipulation techniques. Other models in our comparative analysis were also trained on multiple different datasets. The most complete and detailed results were provided in a work on XceptionNet with video transformers (XNet-ViT) by Khan and Dai [30]. We also compare our results with the following works: Xception by Rossler [46], Audio-visual deepfake detection method (AVA-SN) by Mittal et al. [40], Co-motion (Wang G., Zhou J. and Wu Y [56]), S-MIL [35], STIL [17], I-I-SIM by Zhihao Gu, Yang Chen and Taiping Yao [18], [19], V4D [64], TEI [39] and DSANet [59], method based on 3D decomposition (FD2Net) by Zhu et al. [66], Face X-Ray method by Li et al. [34], EffNetB4Att (EfficientNetB4 with Attention) by Bonettini [7], a method employing recurrent neural network (Conv-LSTM) by Guera et al. [20], MesoNet by Afchar et al. [1] and CLRNet by Shahroz Tariq [53].

	\subsection{Comparing with state-of-the-art}
	We conducted a comparative analysis with state-of-the-art models that were trained on multiple datasets employing several different scenarios for evaluating our model accuracy.
	
	In the first scenario we have evaluated the performance of our model trained on a combined dataset including FaceSwap, Deepfake, Face2Face and Neural Texture from FF++ dataset, DFD and DFDC dataset. The datasets were split into train and test subsets with 80:20 ratio with a 5-fold cross-validation, and the performance of the model was evaluated on each deepfake generation algorithm separately. Our model outperforms the approaches by Khal [30], model Xception [46], MesoNet by Afchar [1] and CLRNet by Shahroz Tariq [53] in terms of F-measure across all datasets. The results are presented in Table~\ref{table:table1}.

	In the second scenario we have evaluated our model when trained on a separate dataset with the performance measured on the same dataset it was trained on.  The best results on FaceForensics++ were demonstrated by Intra-Inter-SIM (I-I-SIM) by Zhihao Gu, Yang Chen and Taiping Yao [18], [19], therefore we took their results for comparison. Following their setup, we trained our model on FaceForensics++ dataset and measured the model performance for each manipulation algorithm (DF, F2F, FS, NT) and each video quality (HQ, LQ) separately. The results in terms of accuracy are provided in Table~\ref{table:table2}. Similarly, the model performance was evaluated on FaceForensics++, DFD, DFDC and Celeb-DF datasets. The results in terms of accuracy are provided in Table~\ref{table:table3}.
	
	As can be seen from Table~\ref{table:table2} overall our model outperforms state-of-the-art except for NT subset of FF++ dataset, where the results are slightly lower. We believe this happens because in NT subset only a mouth area is affected by manipulation, whereas the model evaluates many different features of a face and not just the mouth area. Zhihao Gu et al. [19] analyze the mouth area separately therefore their performance on NT subset is higher.

	\begin{table}[!htbp]
		\caption{Comparison with other deepfake detection methods on FF+, DFD, Celeb-DF and DFDC datasets (results are presented in terms of accuracy).}
		\label{table:table3}
		\centering
		\begin{tabular}{@{}lllll@{}}
			\toprule
			\shortstack[l]{Method} & \shortstack[l]{FF+} & \shortstack[l]{DFD} & \shortstack[l]{DFDC} & \shortstack[l]{Celeb-DF} \\
			\midrule
			Xception & 0.9573 & 0.8807 & 0.8560 & 0.9944 \\
			AVA-SN & - & - & 0.8440 & -\\
			FD2Net & 0.9961 & 0.8984 & 0.8793 & -\\
			X-Ray & - & 0.9334 & 0.7352 & -\\
			EffNetB4Att & - & 0.8935 & 0.8571 & -\\
			Conv-LSTM & 0.8310 & - & - & -\\
			XNet-ViT & 0.9979 & 0.9928 & 0.9169 & -\\
			S-MIL & - & - & 0.8378 & 0.9923\\
			DSANet & - & - & 0.8867 & 0.9942\\
			STIL & - & - & 0.8980 & 0.9961\\
			I-I-SIM & - & - & 0.9279 & 0.9961\\
			\textbf{Ours (GFF)} & \textbf{0.9984} & \textbf{0.9987} & \textbf{0.9883} & \textbf{0.9994}\\
			\bottomrule
		\end{tabular}
	\end{table}

	Similarly, Table~\ref{table:table3} demonstrate that our model outperforms state-of-the-art on DFD, DFDC, Celeb-DF and all of FF++ datasets (without splitting into separate algorithms). Apart from the measurements provided above in Tables~\ref{table:table1}---~\ref{table:table3}, we have also evaluated the generalizing ability of our model on WildDeepFake dataset since it includes a high diversity of scenes and other conditions and is commonly considered as a challenging dataset for deepfake detection. For this evaluation the model was trained on a combined dataset consisting of FF+, DFD and DFDC datasets (merge learning). We would like to point out that the data from WildDeepFake has not been used in training our model. The comparison was performed with Xception, MesoNet and CLRNet from the work by Shahroz Tariq [53]. The results of the evaluation are provided in Table~\ref{table:table4}.

	\begin{table}[!htbp]
		\caption{Performance comparison with other deepfake detection methods on a previously unseen WildDeepFake datasets in terms of F-measure}
		\label{table:table4}
		\centering
		\begin{tabular}{@{}ll@{}} 
			\toprule
			\shortstack[l]{Method} & F-measure \\
			\midrule
			\shortstack[l]{Conv-LSTM} & 0.5375\\
			\shortstack[l]{DBiRNN} & 0.5197\\
			\shortstack[l]{ShallowNet} & 0.5784\\
			\shortstack[l]{Xception} & 0.7145\\
			\shortstack[l]{MesoNet} & 0.7812\\
			\shortstack[l]{CLRNet} & 0.8495\\
			\shortstack[l]{\textbf{Ours (GFF)}} & \textbf{0.8535} \\
			\bottomrule
		\end{tabular}
	\end{table}

	The experimental results show that our model outperforms current state-of-the-art across different datasets and image manipulation techniques and demonstrate a good ability to generalize to previously unseen deepfake generation algorithms without significantly sacrificing its accuracy.

	\section{Ablation study}
	We verify the effectiveness of the components of the proposed method as follows:
	1)	To see the effect of a CNNblock architecture we have performed experiments with different number of convolution layers (from 1 to 4), types of polling (avg/max) and different kernel sizes (from 1 to 32).
	2)	To see the effect of geometric-fakeness features design we have measures the performance of our approach without the geometric component.
	In order to better demonstrate the influence of the components of our approach in a challenging setup we have selected only videos with multiple faces simultaneously present in a frame. There videos are a subset of FF++ and DFDC datasets (omitting videos with just one face). The resulting dataset contained videos with 2 to 15 faces present simultaneously.
	
	An ablation study on CNNBlock architecture demonstrated that a 2-layer network (Fig.~\ref{fig:mesh6}) performed slightly better than a 1-layer network. Increasing a number of layers further did not yield any improvement. The combination of average and max pooling demonstrated a small improvement over other combinations. And kernel sizes larger than 8 did not yield any improvement over the original setup. The results for a modified CNNBlock with one layer and a kernel sizes from 1 to 4 are provided in Table~\ref{table:table5}.
	
	\begin{table}[!htbp]
		\caption{The results of the ablation study on videos with multiple faces.}
		\label{table:table5}
		\centering
		\begin{tabular}{@{}lll@{}}
			\toprule
			\shortstack[l]{Method} & ROC-AUC & F-measure \\
			\midrule
			\shortstack[l]{S-MIL-T} & 0.9131  & 0.8643 \\
			\shortstack[l]{STIL} & 0.9185 & 0.8695 \\
			\shortstack[l]{I-I-SIM}  & 0.9462 & 0.8991 \\
			\shortstack[l]{GFF} & \textbf{0.9639} & \textbf{0.9466} \\
			\shortstack[l]{Modified CNNBlock} &  0.9591 & 0.9415 \\
			\shortstack[l]{Fakeness features only} & 0.9426 & 0.9002  \\
			\bottomrule
		\end{tabular}
	\end{table}
	
	For an ablation study on geometric-fakeness features we have trained our network without computing geometric features using only fakeness features. The results of this approach were significantly lower than in the original setup and are provided in Table~\ref{table:table5}.

	As can be seen from the results, geometric features have a major influence on the performance of our method. Without them, we do not achieve SotA results. Whereas with all the components combined our approach (Figs.~\ref{fig:mesh5}---~\ref{fig:mesh6}) outperforms other SotA approaches.

	\section{Conclusions}
	In this work we have proposed a method for deepfake detection that demonstrates a state-of-the-art performance and generalizes well across different datasets and image manipulation techniques without sacrificing its accuracy. We proposed an approach to calculate GFF thus considering both temporal and geometric information about each separate face in a video. This allows the model to perform well even in difficult conditions where multiple faces are present in a video simultaneously. Also our method has a good generalizing ability and performs well on previously unseen deepfake generation algorithms. From a practical standpoint, our method allows to improve the security of facial biometric, brand monitoring or online video conferencing solutions against attacks employing deepfakes since it is designed to process multiple faces simultaneously.

	\section*{References}

	{
		\small

		[1] Afchar, D.; Nozick, V.; Yamagishi, J.; Echizen, I. 2018. MesoNet: a compact facial video forgery detection network. In Proceedings of the IEEE International Workshop on Information Forensics and Security, 1–7.

		[2] Agarwal, S.; Farid, H.; Gu, Y.; He, M.; Nagano, K.; Li, H. 2019 Protecting world leaders against deep fakes. In Proceedings of the IEEE Conference on Computer Vision and Pattern Recognition, 38-45.

		[3] Aliev, A. 2020. Elon Musk joined our Zoom call Avatarify. https://www.youtube.com/watch?v=lONuXGNqLO0. Accessed: 2023-10-11.

		[4] Avatarify Python. 2021. Photorealistic avatars for video-conferencing. https://github.com/alievk/avatarify-python, 10 Aug 2021. Attribution-NonCommercial 4.0 International License. Accessed: 2023-11-09.

		[5] Bappy, J.H.; Simons, C.; Nataraj, L.; Manjunath, B.S.; Roy-Chowdhury, A.K. 2019. Hybrid LSTM and encoder-decoder architecture for detection of image forgeries. arXiv:1903.02495v1.

		[6] Bayar, B. and Stamm, M.C. 2016. A deep learning approach to universal image manipulation detection using a new convolutional layer. In Proceedings of the 4th ACM Workshop on Information Hiding and Multimedia Security, 5-10.

		[7] Bonettini, N.; Cannas, E.D.; Mandelli, S.; Bondi, L.; Bestagini, P.; Tubaro, S. 2021. Video face manipulation detection through ensemble of CNNs. In Proceedings of the 25th International Conference on Pattern Recognition, 5012–5019.

		[8] Chen, S.; Yao, T.; Chen, Y.; Ding, S.; Li, J.; and Ji, R. 2021. Local Relation Learning for Face Forgery Detection. In Proceedings of the AAAI Conference on Artificial Intelligence, 1081–1088.

		[9] Liang Chen, Yong Zhang, Yibing Song, Lingqiao Liu, Jue Wang. 2020. Self-Supervised Learning of Adversarial Example: Towards Good Generalizations for Deepfake Detection. Proceedings of the IEEE/CVF Conference on Computer Vision and Pattern Recognition (CVPR), 18710-18719.

		[10] DFCA - Deep Fake Challenge Association. 2021. Tinkov's doppelganger invites to a fake site. https://deepfakechallenge.com/gb/2021/09/16/11906, 16 September 2021. Accessed: 2023-11-11.

		[11] DeepFaceLab. 2021. The leading software for creating deepfakes. https://github.com/iperov/DeepFaceLab., 20 November 2021. GPL-3.0 License. Accessed: 2023-11-11.

		[12] Dolhansky, B.; Howes, R.; Pflaum, B.; Baram, N.; and Ferrer, C. C. 2019. The deepfake detection challenge (dfdc) preview dataset. arXiv preprint arXiv:1910.08854.

		[13] Xiaoyi Dong, Jianmin Bao, Dongdong Chen, Ting Zhang, Weiming Zhang, Nenghai Yu, Dong Chen, Fang Wen, Baining Guo. 2022. Protecting Celebrities From DeepFake With Identity Consistency Transformer. In Proceedings of the IEEE Conference on Computer Vision and Pattern Recognition.

		[14] Du, M.; Pentyala, S.K.; Li, Y.; Hu, X. 2019. Towards generalizable forgery detection with locality-aware autoencoder. ArXiv: 1909.05999.

		[15] Durall, R.; Keuper, M.; Keuper, J. 2020. Watch your up-convolution: CNN based generative deep neural networks are failing to reproduce spectral distributions. In Proceedings of the IEEE Conference on Computer Vision and Pattern Recognition, 7887–7896.

		[16] Google AI Blog. 2019. Contributing Data to Deepfake Detection Research. https://ai.googleblog.com/2019/09/contributing-data-to-deepfake-detection.html, 24 September 2019. Accessed: 2023-05-05.

		[17] Gu, Z.; Chen, Y.; Yao, T.; Ding, S.; Li, J.; Huang, F.; and Ma, L. 2021. Spatiotemporal Inconsistency Learning for DeepFake Video Detection. arXiv:2109.01860.

		[18] Gu, Z.; Chen, Y.; Yao T.; Ding, S.; Li, J.; Ma, L. 2022. Delving into the Local: Dynamic Inconsistency Learning for DeepFake Vide Detection. In Proceedings of the Thirty-Sixth AAAI Conference on Artificial Intelligence, 744-752.

		[19] Gu, Z.; Yao T.; Chen, Y.; Yi, R; Ding, S.; Ma, L. 2022. Region-Aware Temporal Inconsistency Learning for DeepFake Video Detection. In Proceedings of the International Joint Conferences on Artificial Intelligence Organization, IJCAI-2022, 920-926.

		[20] Guera, D. and Delp, E.J. 2018. Deepfake video detection using recurrent neural networks. In Proceedings of the 15th IEEE International Conference on Advanced Video and Signal Based Surveillance,1-6.

		[21] Haliassos, A.; Vougioukas, K.; Petridis, S.; and Pantic, M. 2021. Lips Don’t Lie: A Generalisable and Robust Approach To Face Forgery Detection. In Proceedings of the IEEE Conference on Computer Vision and Pattern Recognition, 5039–5049.

		[22] He, K.; Zhang, X.; Ren, S.; and Sun, J. 2016. Deep residual learning for image recognition. In Proceedings of the IEEE conference on computer vision and pattern recognition, 770–778.

		[23] Heun, D. 2021. Facial recognition tech is catching on with banks. American Banker. https://www.americanbanker.com/news/facial-recognition-tech-is-catching-on-with-banks. 13 September 2021. Accessed: 2023-10-09.

		[24] Hsu, C.C.; Lee, C.Y.; Zhuang, Y.X. 2018. Learning to detect fake face images in the wild. In Proceedings of the International Symposium on Computer, Consumer and Control, 388–391.

		[25] Hu, J.; Shen, L.; Albanie, S.; Sun, G.; Wu, E. 2020. Squeeze-and-excitation networks. In Proceedings of the IEEE Transactions on Pattern Analysis and Machine Intelligence 42, 2011–2023.

		[26] Islam, A.; Long, C.; Basharat, A.; Hoogs, A.J. 2020. DOA-GAN: Dual-order attentive generative adversarial network for image copy-move forgery detection and localization. In Proceedings of the IEEE Conference on Computer Vision and Pattern Recognition, 4675–4684.

		[27] Jensen, G. 2021. How Banks and Financial Institutions Use Face Recognition to Protect People, Property, and Assets. The BriefCam. https://www.briefcam.com/resources/blog/how-banks-and-financial-institutionsuse-face-recognition-to-protect-people-property-and-assets/, 14 January 2021. Accessed: 2023-10-09.

		[28] Kim, M., Tariq, S., \& Woo, S. S. 2021. Cored: Generalizing fake media detection with continual representation using distillation. In Proceedings of the 29th ACM International Conference on Multimedia, 337-346.

		[29] Kim, M., Tariq, S., and Woo, S. S. 2021. Fretal: Generalizing deepfake detection using knowledge distillation and representation learning. In Proceedings of the IEEE/CVF conference on computer vision and pattern recognition, 1001-1012.

		[30] Khan, S.A. and Dai, H. 2021. Video transformer for deepfake detection with incremental learning. In Proceedings of the of the 29th ACM International Conference on Multimedia, 1821-1828.

		[31] Kowalski, M. 2021. FaceSwap: 3D face swapping implemented in Python. https://github.com/MarekKowalski/FaceSwap, 13 April 2021. MIT license. Accessed: 2023-11-11.

		[32] Le, T.N.; Nguyen, H.H.; Yamagishi, J.; Echizen, I. 2021.  OpenForensics: Large-scale challenging dataset for multi-face forgery detection and segmentation in-the-wild. In Proceedings of the IEEE International Conference on Computer Vision, 10097–10107.

		[33] Lee, S., Tariq, S., Kim, J. and Woo, S. S. 2021. Tar: Generalized forensic framework to detect deepfakes using weakly supervised learning. In IFIP International Conference on ICT Systems Security and Privacy Protection, 351-366. Springer, Cham.

		[34] Li, L.; Bao, J.; Zhang, T.; Yang, H.; Chen, D.; Wen, F.; Guo, B. 2020. Face X-Ray for more general face forgery detection In Proceedings of the IEEE Conference on Computer Vision and Pattern Recognition, 5000–5009.

		[35] Li, X.; Lang, Y.; Chen, Y.; Mao, X.; He, Y.; Wang, S.; Xue, H.; and Lu, Q. 2020. Sharp multiple instance learning for deepfake video detection. In Proceedings of the 28th ACM international conference on multimedia, 1864–1872.

		[36] Li, Y. and Lyu, S. 2019. Exposing DeepFake Videos By Detecting Face Warping Artifacts. In Proceedings of the IEEE Conference on Computer Vision and Pattern Recognition Workshops, 46–52.

		[37] Li, Y.; Yang, X.; Sun, P.; Qi, H.; and Lyu, S. 2020. Celeb-DF: A large-scale challenging dataset for deepfake forensics. In Proceedings of the IEEE Conference on Computer Vision and Pattern Recognition, 3207–3216.

		[38] Lithuanian radio and Television, LRT English Newsletter. 2021. Imposter used deepfake to dupe Baltic MPs, impersonate Navalny associate. https://www.lrt.lt/en/news-in-english/19/1393935/imposter-used-deepfake-to-dupe-baltic-mps-impersonate-navalny-associate, 23 April 2021. Accessed: 2023-10-09.

		[39] Liu, Z.; Luo, D.; Wang, Y.; Wang, L.; Tai, Y.; Wang, C.; Li, J.; Huang, F.; and Lu, T. 2020. Teinet: Towards an efficient architecture for video recognition. In Proceedings of the AAAI Conference on Artificial Intelligence, 11669–11676.

		[40] Mittal, T.; Bhattacharya, U.; Chandra, R.; Bera, A.; Manocha, D. 2020. Emotions don't lie: An audio-visual deepfake detection method using affective cues. In Proceedings of the of the 28th ACM International Conference on Multimedia, 2823-2832.

		[41] Nirkin, Y.; Wolf, L.; Keller, Y.; Hassner, T. 2020. Deepfake [ et al., 20] detection based on the discrepancy between the face and its context. ArXiv: 2008.12262.

		[42] Pashaeva, Y. 2021. Scammers Are Using Deepfake Videos Now. Online magazine "Slate". https://slate.com/technology/2021/09/deepfake-video-scams. html, 13 September 2021. Accessed: 2023-10-09.

		[43] Perov, I.; Gao, D.; Chervoniy, N.; Liu, K.; Marangonda, S.; Ume, C.; Dpfks, M.; Facenheim, C.S.; RP, L.; Jiang, J.; Zhang, S.; Wu, P.; Zhou, B.; Zhang, W. 2020. DeepFaceLab: Integrated, flexible and extensible face-swapping framework. arXiv preprint arXiv:2005.05535.

		[44] Qian, Y.; Yin, G.; Sheng, L.; Chen, Z.; and Shao, J. 2020. Thinking in frequency: Face forgery detection by mining frequency-aware clues. In European Conference on Computer Vision, 86–103. Springer.

		[45] Reuters. U.S. President Joe Biden delivers the State of the Union address to a joint session of Congress at the U.S. Capitol in Washington, DC, U.S, March 1, 2022. Saul Loeb/Pool via REUTERS.

		[46] Rossler, A.; Cozzolino, D.; Verdoliva, L.; Riess, C.; Thies, J.; and Nießner, M. 2019. Faceforensics++: Learning to detect manipulated facial images. In Proceedings of the IEEE International Conference on Computer Vision, 1–11.

		[47] Sabir, E.; Cheng, J.; Jaiswal, A.; AbdAlmageed, W.; Masi, I.; and Natarajan, P. 2019. Recurrent convolutional strategies for face manipulation detection in videos. Interfaces (GUI), 3(1): 80–87.

		[48] Salloum, R; Ren, Y; C-C. Jay Kuo. 2017. Image Splicing Localization Using A Multi-Task Fully Convolutional Network (MFCN). arXiv preprint arXiv:1709.02016.

		[49] Schroff, F.; Kalenichenko, D.; Philbin, J. 2015. FaceNet: A Unified Embedding for Face Recognition and Clustering. arXiv preprint arXiv:1503.03832.

		[50] Kaede Shiohara, Toshihiko Yamasaki. 2022. Detecting Deepfakes With Self-Blended Images.Proceedings of the IEEE/CVF Conference on Computer Vision and Pattern Recognition (CVPR), 18720-18729.

		[51] Stehouwer, J.; Dang, H.; Liu, F.; Liu, X.; Jain, A.K. 2020. On the detection of digital face manipulation. In Proceedings of the IEEE Conference on Computer Vision and Pattern Recognition, 5780–5789.

		[52] Tan, M. and Le, Q.V. 2019. EfficientNet: Rethinking model scaling for convolutional neural networks. ArXiv: 1905.11946.

		[53] Tariq, S., Lee, S., \& Woo, S. 2021. One detector to rule them all: Towards a general deepfake attack detection framework. In Proceedings of the Web Conference 2021, 3625-3637.

		[54] The Guardian. 2021. European MPs targeted by deepfake video calls imitating Russian opposition. https://www.theguardian.com/world/2021/apr/22/ europeanmps-targeted-by-deepfake-video-calls-imitating-russian-opposition, 22 April 2021. Accessed: 2023-11-11.

		[55] The OBS Project. OBS Studio - Stream with Power and Ease. Open Broadcaster Software. https://www.obsstudio.net. Accessed: 2023-11-11.

		[56] Wang, G.; Zhou, J.; and Wu, Y. 2020. Exposing Deepfaked Videos by Anomalous Co-motion Pattern Detection. In arXiv preprint arXiv:2008.04848.

		[57] Xueyu Wang, Jiajun Huang, Siqi Ma, Surya Nepal, Chang Xu. 2022. DeepFake Disrupter: The Detector of DeepFake Is My Friend. Proceedings of the IEEE/CVF Conference on Computer Vision and Pattern Recognition (CVPR), 14920-14929.

		[58] Wu, Y.; AbdAlmageed, W.; Natarajan, P. 2019. Mantra-net: Manipulation tracing network for detection and localization of image forgeries with anomalous features. In Proceedings of the IEEE Conference on Computer Vision and Pattern Recognition, 9535–9544.

		[59] Wu, W.; Zhao, Y.; Xu, Y.; Tan, X.; He, D.; Zou, Z.; Ye, J.; Li, Y.; Yao, M.; Dong, Z.; et al. 2021. DSANet: Dynamic Segment Aggregation Network for Video-Level Representation Learning. arXiv preprint arXiv:2105.12085.

		[60] Yang Z, Liang J, Xu Y, Zhang XY, He R. Masked Relation Learning for DeepFake Detection. IEEE Transactions on Information Forensics and Security (TIFS). 18:1696–1708, 2023.

		[61] Yuting Xu, Jian Liang, Gengyun Jia, Ziming Yang, Yanhao Zhang, Ran He. TALL: Thumbnail Layout for Deepfake Video Detection. IEEE/CVF International Conference on Computer Vision (ICCV), 2023.

		[62] Yuting Xu, Jian Liang, Lijun Sheng, Xiao-Yu Zhang. Towards Generalizable Deepfake Video Detection with Thumbnail Layout and Graph Reasoning. IJCV 2024: International Journal of Computer Vision, 2024.

		[63] Zhang, K.; Zhang, Z.; Li, Z.; and Qiao, Y. 2016. Joint Face Detection and Alignment Using Multitask Cascaded Convolutional Networks. IEEE Signal Processing Letters, 23(10): 1499–1503.

		[64] Zhang, S.; Guo, S.; Huang, W.; Scott, M. R.; and Wang, L. 2020. V4d: 4d convolutional neural networks for video-level representation learning. arXiv preprint arXiv:2002.07442.

		[65] Zhang, X.; Karaman, S.; Chang, S.F. 2019. Detecting and simulating artifacts in GAN fake images. In Proceedings of the IEEE International Workshop on Information Forensics and Security.

		[66] Zhu, X.; Wang, H.; Fei, H.; Lei, Z.; Li, S. 2021. Face forgery detection by 3d decomposition. In Proceedings of the IEEE Conference on Computer Vision and Pattern Recognition, 2928-2938.

		[67] Zi, B.; Chang, M.; Chen, J.; Ma, X.; and Jiang, Y.G. 2020. WildDeepFake: A challenging real-world dataset for deepfake detection. In Proceedings of the 28th ACM International Conference on Multimedia, 2382–2390.

	}
	

\end{document}